\documentclass[sigconf]{acmart} 
\AtBeginDocument{%
  }

\setcopyright{acmlicensed} 
\copyrightyear{2026} 
\acmYear{2026} 
\acmDOI{XXXXXXX.XXXXXXX} 
\acmConference[AISec '26]{19th ACM Workshop on Artificial Intelligence and Security}{November 15, 2026}{The Hague, Netherlands}  
\acmISBN{978-1-4503-XXXX-X/2018/06}

\usepackage{tikz}
\usepackage{url}
\usetikzlibrary{
  arrows.meta,
  positioning,
  calc,
  fit,
  backgrounds,
  shapes.geometric,
}

\usepackage{enumitem}
\usepackage{float}
\usepackage{algorithm}
\usepackage{algpseudocode}

\newcommand{\yes}{{\color{green!50!black}\checkmark}}
\newcommand{\no}{{\color{red!70!black}$\times$}}
\newcommand{\partialmark}{{\color{orange!70!black}$\sim$}}

\hypersetup{colorlinks=true,linkcolor=blue!60!black,citecolor=blue!60!black,urlcolor=blue!60!black}
\begin{document}



\title{Share No More Than the Request Requires: 
Federated Disclosure for Perspective-Aware AI}



\author{%
  Sourena Khanzadeh$^{1,2,*}$ \quad
  Daniel Platnick$^{2}$ \quad
  Marjan Alirezaie$^{1,2}$ \quad
  Hossein Rahnama$^{1,2,3}$%
}
\affiliation{%
  \institution{$^{1}$Toronto Metropolitan University \quad
               $^{2}$Flybits Labs \quad
               $^{3}$MIT Media Lab}
  \city{Toronto}
  \country{Canada}%
}
\email{sourena.khanzadeh@torontomu.ca}
\email{daniel.platnick@flybits.com}
\email{malirezaie@torontomu.ca}
\email{rahanma@mit.edu}

\authorsaddresses{$^{*}$Corresponding author: Sourena Khanzadeh
  (sourena.khanzadeh@torontomu.ca).}







\renewcommand{\shortauthors}{Khanzadeh et al.}

\begin{abstract} 


Modern AI systems bring societal risks such as mass surveillance, extreme concentrations of power, and loss of user autonomy---calling into question a model where third-parties collect and control massive amounts of user data.
Users require a sovereign system to securely own, govern, and disclose their context while remaining compliant across regulated domains with strict provenance, interpretability, and policy adherence.
Perspective-aware AI approaches this by transforming a user's aggregated personal data into a structured identity model called a \emph{Chronicle}: a temporal knowledge graph that represents and grows with the user.
Chronicles support the secure disclosure of context across federated networks.
A Chronicle holder may expose a queryable, authorized view that a third-party agent may consult without centralizing anyone's data. 
This paper explores the problem of minimum-necessary disclosure across domain boundaries: when a requester's agent queries a Chronicle, how can the system constrain its response to release only what the requester's relationship, stated purpose, and specific task require?
We propose \textbf{Provenance Preserving Chronicles} (PPC), a federated protocol that compiles each holder's Chronicle into a compact \emph{authorized evidence subgraph} governed by one rule:
\emph{share no more than the request requires}. 
Holders keep local sovereignty; an access controller projects relationship-aware views over domain-expert ontologies;
and a two-phase flow returns provenance-linked text first, releasing raw artifacts only after explicit holder approval. 
We frame the problem, map gaps in blockchain, P2P, and holder-sovereign designs, define the core constructs, and sketch the protocol with an explicit threat model.
\end{abstract}

\ccsdesc[500]{Security and privacy~Software and application security}
\ccsdesc[500]{Social and professional topics~Privacy policies}
\ccsdesc[500]{Social and professional topics~Digital rights management}


\keywords{User sovereignty, digital identity, selective disclosure, data governance, federated networks, perspective-aware AI} 


\maketitle

\section{Introduction}
\label{sec:intro}

Third-party platforms aggregate and control ever-larger troves of personal data, leaving users owning less and less of their digital identity~\cite{info12110465}.
Repeated surveillance disclosures and security breaches have increased demand for architectures in which users own and control what they
share~\cite{fernandez2013security,cavoukian2009privacy}, preserving sovereignty over their data. 
Bitcoin demonstrated that trusted, auditable coordination is possible without a central authority, using a decentralized network of peers accompanied by a public ledger~\cite{nakamoto2008bitcoin} creating a trusted environment for entities making transactions. 
MedRec applied that pattern to healthcare records: a decentralized personal data management system that turns a blockchain into an automated access-control manager---one that does not require trust in a third party---whose transactions carry instructions to store, query, and share data rather than strictly financial transfers~\cite{azaria2016medrec}.
The direction is right; the substrate is incomplete for autonomous agent-native identity exchange.

The premise that individuals should hold and \emph{selectively} share portable,
machine-verifiable identity context is not unique to any one group: it underpins self-sovereign identity~\cite{allen2016ssi}, Solid personal data pods~\cite{bernerslee2018solid}, and the W3C Decentralized Identifier and
Verifiable Credential stack~\cite{sporny2022did,sporny2025vc}.
Perspective-Aware AI (PAi) advances one such form, treating identity as something you can share, not something a platform owns~\cite{alirezaie2025perspective}.

PAi builds this identity from the bottom up. 
A user opts in by connecting their data channels (e.g., third-party apps) and contributing their data artifacts such as images, audio, text.
Artifacts are then mapped to situation-level graphs describing distinct \emph{situations} the user has experienced.
For provenance, data artifacts are stored with pointers to their corresponding situation graphs.
Compiling the user's full set of situation graphs across time yields a \emph{Chronicle}~\cite{alirezaie2025perspective}: a longitudinal knowledge graph of how a person or organization shows up across contexts. 
Chronicles already power agents in deployed settings such as social simulation and extended reality ~\cite{platnick2025perspective,platnick2025id}.
The next step is to make Chronicles securely shareable as a user-owned, digital identity that a third-party agent may temporarily query as a source of personal
context---without copying or owning the underlying identity data. 
On a federated network, this lets one node's agent permissibly consult another user's
Chronicle without centralizing anyone's data or routing disclosure through a global ledger.

That portability breaks down at the disclosure boundary. 
\emph{This paper addresses the problem of selective disclosure for PAi Chronicles: given a requester, a relationship, a purpose, and a query, how can a holder release the smallest authorized subgraph sufficient to answer the request?} 
In a doctor--patient relationship, each side holds a medical Chronicle---conditions, treatments, outcomes---and agents on both sides could exchange clinical perspective to inform care. 
A cardiologist checking drug interactions does not need psychiatric session notes; an insurance auditor does not need the same slice as a treating physician.
To minimize disclosure in the controlled exchange of identity context between two users, each release should be scoped to the requester's \emph{relationship}, the stated \emph{purpose}, and the \emph{information need}---not shipped as an all-or-nothing export. 
The problem is not how to retrieve from a graph; it is how to \emph{share Chronicle evidence under minimum-necessary constraints by a dynamic access control} when holders, policies, and domains differ.

To our knowledge, no existing framework jointly answers three questions at exchange time: \emph{why} does the requester need this data, \emph{what relationship} do they hold to the subject, and \emph{what is the smallest authorized fragment} sufficient for the task? 
Blockchain access managers log who may touch which records but do not compile predicate-labeled, temporally ordered evidence subgraphs per request; holder-sovereign platforms such as Solid~\cite{bernerslee2018solid} and P2P overlays distribute custody without exchange-time minimization (\S\ref{sec:gap}).

Agent-mediated information exchange creates a structural tension between useful personalization and the holder’s control over sensitive records. We propose \textbf{Provenance Preserving Chronicles} (PPC), a federated protocol for selective, auditable disclosure across autonomous agents. PPC represents a holder’s records as locally governed Chronicles and uses domain ontologies, relationship-aware policies, and dynamic subgraph projection to construct request-specific evidence views. Its two-phase interaction model first produces provenance-linked textual responses, while access to underlying artifacts requires a separate, explicit authorization step. The design distinguishes protocol-level access-control guarantees from risks introduced after disclosure, including downstream inference, prompt injection, and the irreversible retention of information by LLM-based consumers. We formalize the system’s threat model and illustrate its operation through medical and litigation scenarios. These case studies show how a common protocol can support multiple regulated domains by varying ontologies, participant roles, and disclosure policies without centralizing the underlying data.



\section{Background}
\label{sec:prelim}

We build on two PAi constructs: Situation Graphs (\S\ref{sec:sg}) and
Chronicles (\S\ref{sec:chronicles}).

\subsection{Situation Graphs}
\label{sec:sg}

A Situation Graph (SG) snapshots one situated moment as a predicate-labeled
knowledge graph. By \emph{predicate-labeled} we mean every edge carries a
semantic type---a predicate from a domain ontology---so the graph encodes
\emph{typed} relations rather than untyped connectivity, and policy and
retrieval reason over these types. A central \emph{situation node} anchors
context---participant, location, time, weather, emotion, activity, and related
entities. Raw identity artifacts become machine-readable structure that can be
sequenced over time (Figure~\ref{fig:compact-chronicle}). 

\subsection{Chronicles}
\label{sec:chronicles}

At time $t_i$, a Situation Graph is a knowledge graph
\[
G_{t_i} = (V_{t_i}, E_{t_i}),
\]
where $V_{t_i}$ holds entities, attributes, and contextual objects, and
\[
E_{t_i} \subseteq V_{t_i} \times \mathcal{P} \times V_{t_i}
\]
holds directed predicate-labeled edges. Each edge is a semantic triple
$(s,p,o)$---subject, predicate $p \in \mathcal{P}$, object. A distinguished
situation node~$s_{t_i}$ anchors the moment, but edges need not radiate from it:
contextual entities also link directly (e.g.,
$(\mathit{lamotrigine},\mathit{hasDosage},\\ 200\mathit{mg})$), so $G_{t_i}$ is a
general typed graph, not a star centered on~$s_{t_i}$.

A Chronicle for entity~$u$ is the time-ordered sequence of those snapshots:
\[
\mathcal{C}_u = \langle G_{t_1}, G_{t_2}, \ldots, G_{t_n} \rangle,
\]
with $t_1 < t_2 < \cdots < t_n$. Each $G_{t_i}$ is how~$u$ showed up at
time~$t_i$; together they track context, behavior, preferences, and perspective
as they evolve.

The sequence is the backbone, but a Chronicle is more than a list of snapshots:
each $G_{t_i}$ carries \emph{temporal anchors} (event and ingestion timestamps)
and \emph{provenance} linking its triples to the source artifacts they were
extracted from; \emph{cross-situation links} join co-referent entities across
snapshots; and the consolidated view $G_{\mathcal{C}_u}$ (below), plus any
\emph{inferred patterns} (habits, trends) mined over it, are part of the
Chronicle too. These are first-class, though the formalism foregrounds the
sequence and its consolidation.

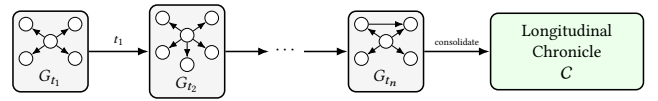
\begin{figure}[h]
\centering
\resizebox{\linewidth}{!}{
\begin{tikzpicture}[
    sg/.style={
        draw, rounded corners, thick,
        minimum width=.2cm, minimum height=.25cm,
        align=center, fill=gray!8
    },
    chronicle/.style={
        draw, rounded corners, thick,
        minimum width=2.8cm, minimum height=1.4cm,
        align=center, fill=green!8
    },
    nodept/.style={
        circle, draw, fill=white, inner sep=1.2pt
    },
    arrow/.style={-{Latex[length=2mm]}, thick}
]
\node[sg] (g1) {
\begin{tikzpicture}[scale=0.45]
    \node[nodept] (s) at (0,0) {};
    \node[nodept] (p) at (-1,0.7) {};
    \node[nodept] (l) at (1,0.7) {};
    \node[nodept] (a) at (-1,-0.7) {};
    \node[nodept] (e) at (1,-0.7) {};
    \draw[-{Latex[length=1.5mm]}] (s)--(p);
    \draw[-{Latex[length=1.5mm]}] (s)--(l);
    \draw[-{Latex[length=1.5mm]}] (s)--(a);
    \draw[-{Latex[length=1.5mm]}] (s)--(e);
\end{tikzpicture}\\[-1mm]
$G_{t_1}$
};
\node[sg, right=1.1cm of g1] (g2) {
\begin{tikzpicture}[scale=0.45]
    \node[nodept] (s) at (0,0) {};
    \node[nodept] (p) at (-1,0.7) {};
    \node[nodept] (l) at (1,0.7) {};
    \node[nodept] (a) at (-1,-0.7) {};
    \node[nodept] (e) at (1,-0.7) {};
    \node[nodept] (w) at (0,-1.2) {};
    \draw[-{Latex[length=1.5mm]}] (s)--(p);
    \draw[-{Latex[length=1.5mm]}] (s)--(l);
    \draw[-{Latex[length=1.5mm]}] (s)--(a);
    \draw[-{Latex[length=1.5mm]}] (s)--(e);
    \draw[-{Latex[length=1.5mm]}] (s)--(w);
\end{tikzpicture}\\[-1mm]
$G_{t_2}$
};
\node[right=0.8cm of g2] (dots) {$\cdots$};
\node[sg, right=0.8cm of dots] (gn) {
\begin{tikzpicture}[scale=0.45]
    \node[nodept] (s) at (0,0) {};
    \node[nodept] (p) at (-1,0.7) {};
    \node[nodept] (l) at (1,0.7) {};
    \node[nodept] (a) at (-1,-0.7) {};
    \node[nodept] (e) at (1,-0.7) {};
    \draw[-{Latex[length=1.5mm]}] (s)--(p);
    \draw[-{Latex[length=1.5mm]}] (s)--(l);
    \draw[-{Latex[length=1.5mm]}] (s)--(a);
    \draw[-{Latex[length=1.5mm]}] (s)--(e);
    \draw[-{Latex[length=1.5mm]}] (p)--(l);
\end{tikzpicture}\\[-1mm]
$G_{t_n}$
};
\node[chronicle, right=1.2cm of gn] (c) {
Longitudinal\\Chronicle\\$\mathcal{C}$
};
\draw[arrow] (g1) -- node[above, font=\scriptsize] {$t_1$} (g2);
\draw[arrow] (g2) -- (dots);
\draw[arrow] (dots) -- (gn);
\draw[arrow] (gn) -- node[above, font=\tiny] {consolidate} (c);
\end{tikzpicture}
}
\caption{A Chronicle stacks Situation Graph snapshots in time. Each $G_{t_i}$
is a predicate-labeled graph for one situated moment; consolidation yields the
longitudinal Chronicle~$\mathcal{C}$.}
\label{fig:compact-chronicle}
\end{figure}

For retrieval and compilation, we also use a consolidated view:
\[
G_{\mathcal{C}_u} = \bigoplus_{i=1}^{n} G_{t_i},
\]
where $\bigoplus$ is \emph{not} a plain set union but a
\emph{provenance-preserving merge with entity alignment}: co-referent nodes are
unified across snapshots, every node and edge keeps its time index~$t_i$ and a
provenance pointer to its source Situation Graph and artifact, and duplicate
triples are coalesced with their temporal annotations preserved. The result
$G_{\mathcal{C}_u} = (V_{\mathcal{C}_u}, E_{\mathcal{C}_u})$ is a single
temporal, provenance-annotated identity graph---not a bag of disconnected
snapshots (Figure~\ref{fig:compact-chronicle}).

\section{Related Work}
\label{sec:gap}

This section discusses the existing literature, approaches, and gaps surrounding user-sovereign identity exchange and personal context management.

\subsection{Blockchain and Confidential Smart-Contract Networks}

Blockchain systems enforce decentralized trust through replicated global state and consensus. 
MedRec~\cite{azaria2016medrec} applied this to medical-record access management: providers log permissions on-chain while patients retain copies locally. 
Secret Network~\cite{secretnetwork2020graypaper,li2022teetcsc} and related Trusted Execution Environment (TEE)-assisted platforms (Oasis, Phala) push further with \emph{programmable privacy}---CosmWasm contracts executing inside SGX enclaves so contract inputs, outputs, and state stay encrypted in use~\cite{zyskind2015enigma}. 
Li et al.~\cite{li2022teetcsc} systematize these designs as layer-one and layer-two confidential smart-contract stacks.

For Chronicle disclosure, the blockchain model is a poor fit. 
Validators replicate ledger state across the network; disclosure becomes \emph{contract
execution} over globally agreed state, not \emph{holder-local compilation} of a
minimum authorized subgraph. Relationship, purpose, and predicate-level policy
must be encoded ad hoc per application; there is no native notion of
Situation-Graph structure or exchange-time minimization across sovereign holders.
Security analyses reinforce the mismatch: Jean-Louis et al.~\cite{jeanlouis2024sgxonerated}
show practical privacy breaks on Secret Network from storage access patterns and
state replay; Wilde et al.~\cite{wilde2025forking} document forking and cloning
attacks when TEE state is not tightly coupled to consensus. Chronicle exchange
needs least-authority release between known peers~\cite{warner2010tahoe}, not
global replication with gas-metered contract calls.

\subsection{Federated Holder-Sovereign Platforms}

A second thread keeps data at the edge. Solid~\cite{bernerslee2018solid,mansour2016federation}
stores user artifacts in personal pods with WebID-based access control.
Verifiable-credentials infrastructure~\cite{sporny2025vc,hardjono2019decentralized}
binds identity, role, and institutional affiliation for cross-domain trust
without a central issuer. These models align with holder sovereignty: each party
controls what it stores and who may read it. What they lack is Chronicle
semantics---temporal stacks of predicate-labeled Situation Graphs---and an
exchange-time compiler that returns a compact authorized evidence subgraph
for a stated relationship and purpose.

\subsection{Peer-to-Peer Overlays and Decentralized Storage}

Peer-to-peer overlays distribute content without a global
ledger~\cite{stoica2001chord,maymounkov2002kademlia,benet2014ipfs,trautwein2022design}.
libp2p~\cite{libp2p2024} modularizes discovery, transport, and multiplexing for
application-level federations. Secure Scuttlebutt~\cite{farrell2019ssb} and
similar gossip protocols propagate signed feeds among mutually trusted peers.
OceanStore~\cite{kubiatowicz2000oceanstore} and IPFS-style content addressing
provide durable, location-independent retrieval across untrusted nodes.

These systems solve \emph{where bits live and how they are found}; they do not
decide \emph{which Chronicle fragment a requester may receive}. A P2P overlay can
carry compiled evidence once policy has been applied, but the overlay itself
does not enforce purpose limitation, relationship-scoped predicate access, or
minimum-sufficient subgraph compilation at the holder boundary.

\subsection{Semantic Web Access Control and Policy Languages}

Controlled access to Resource Description Framework (RDF) has been studied for two
decades~\cite{kirrane2017access}. Named graphs attach identity, provenance, and
trust metadata to triple sets, a natural unit for scoping
disclosure~\cite{carroll2005named}; SPARQL~1.1~\cite{harris2013sparql} expresses
filtered views (\texttt{CONSTRUCT}/\texttt{DESCRIBE}, graph patterns,
\texttt{FROM NAMED}); and ontology- and context-based access control attaches
policy to semantic types and request
context~\cite{kagal2003policy,costabello2012context}. Graph-native policy
languages such as ODRL~\cite{iannella2018odrl} and Web Access Control (WAC in
Solid~\cite{bernerslee2018solid}) govern who may operate on which resource, over
triples modeled per RDF~1.1~\cite{cyganiak2014rdf}. This is the right substrate,
and PPC builds on it---predicate types come from a domain-expert ontology
(\S\ref{sec:auth}), and $\mathcal{A}$ can be realized as an ontology-typed,
context-aware policy in this tradition. What none provides is
\emph{exchange-time minimization}: a SPARQL view or WAC rule returns the triples
a requester is \emph{allowed} to see or that a query \emph{matches}, but does not
compile a compact, minimum-necessary evidence subgraph (\S\ref{sec:evidence})
for a relationship and purpose, prune authorized-but-unnecessary structure,
reason over \emph{temporal} Situation-Graph ordering, or coordinate a
provenance-first release across \emph{federated} holders. PPC is the compilation
layer on this substrate, not a replacement.

\subsection{Purpose-, Relationship-, and Norm-Based Access Control}
The two levers PPC pulls at exchange time have long lineages in access
control. Hippocratic databases made purpose a first-class element of a
DBMS: data is tagged with allowed purposes at collection and queries are
filtered against them~\cite{agrawal2002hippocratic}, and Byun and Li
formalized purpose-based access control with purpose hierarchies over
relational data~\cite{byun2008pbac}. Relationship-based access control
(ReBAC) makes the requester--subject relationship, rather than roles or
attributes, the primitive that policy reasons over~\cite{fong2011rebac}.
Contextual integrity frames both as instances of a broader norm:
information flows are appropriate when they conform to context-relative
transmission principles binding sender, recipient, relationship, and
purpose~\cite{nissenbaum2004ci}. Recent work carries these norms to
LLM-based agents: AirGapAgent restricts a conversational agent's working
context to task-necessary user data to resist adversarial
extraction~\cite{bagdasarian2024airgap}, and Ghalebikesabi et al.\
operationalize contextual integrity as an information-flow judgment
inside assistants~\cite{ghalebikesabi2024ci}. These systems act at
prompt or inference time over flat attribute sets. PPC can be read as
the compile-time counterpart for structured identity: the transmission
principle (relationship $r$, purpose $p$) is evaluated by $\mathcal{A}$
before anything crosses the holder boundary, over temporal,
predicate-labeled Chronicle structure, and ``task-necessary'' is
enforced as a subgraph-minimization objective (\S4.2) rather than a
prompt-time heuristic.

\subsection{Positioning: The Closest Alternative Is a Pod Plus a Policy Engine}

We want to be fair about what a ledger is and is not for here. MedRec-style
systems use the blockchain for \emph{permissions and tamper-evident audit}, not
for moving records~\cite{azaria2016medrec}; no serious design routes clinical
payloads or per-disclosure decisions through global consensus. So ``don't run
disclosure over consensus'' is not the real disagreement. A ledger genuinely
buys tamper-evident logging, and PPC could adopt one for exactly that purpose;
what a ledger does \emph{not} provide is any notion of Chronicle structure or
minimum-necessary release, and confidential-contract variants (Secret Network)
add validator/TEE trust and platform-upgrade and forking
risk~\cite{wilde2025forking,jeanlouis2024sgxonerated} without addressing that
gap.

The honest comparison is therefore against the strongest edge-native stack one
could assemble today: \textbf{Solid pods + Verifiable Credentials + a policy
engine}~\cite{bernerslee2018solid,mansour2016federation,sporny2025vc}. That
combination already gets three things right---data stays holder-local, exchange
is ledger-free, and a verifiable presentation can bind a requester's identity,
role, and purpose to an access decision. If the goal were coarse-grained
document access control, it would largely suffice.

Two things are still missing, and they are what PPC contributes. First,
\emph{Chronicle semantics}: pods and policy engines gate documents or RDF
resources, not temporal stacks of predicate-labeled Situation Graphs where the
semantic type of each edge and its position in time are what policy must reason
over. Second, \emph{exchange-time minimization}: a policy engine returns the
resources a requester is \emph{allowed} to see, whereas PPC compiles a
\emph{compact authorized subgraph sufficient for the specific request}
($S^*$)---filtering by relationship and purpose, then pruning toward a
minimum connected evidence set, then shipping provenance-linked text before any
raw artifact. GDPR Art. 5(1)(c) data minimisation and HIPAA's minimum-necessary standard (45 C.F.R. §164.502(b)) ask for exactly this per-request 'smallest sufficient' release \cite{gdpr2016, hipaa_min}, and it is
not something an allow/deny policy layer expresses. PPC is thus complementary to
pods and credentials, not a replacement: it is the compilation step that turns
``what may this requester access?'' into ``what is the least this request
needs?''

Synthesizing these paradigms reveals a distinct architectural gap: temporal, holder-sovereign
Chronicles plus exchange-time, relationship- and purpose-aware evidence
compilation across federated peers. No existing layer supplies that compiler---a
ledger adds tamper-evident audit, a pod-plus-credential stack adds sovereign
storage and access decisions, but neither reasons over Situation-Graph structure
or minimizes what a specific request needs. Chronicles differ from static stores in
three ways---they are \emph{temporal} (ordered Situation Graphs),
\emph{holder-sovereign} (evidence based artifact release is opt-in per holder), and \emph{predicate-rich}
(each edge carries semantic type that policy must respect). Table~\ref{tab:comparison}
maps where existing network systems land on holder sovereignty, ledger-free
exchange, and Chronicle-aware disclosure; P2P overlays and confidential chains
are transport or trust substrates at best, not solutions to minimum-necessary
release.

\begin{figure*}
    \centering
    \resizebox{\linewidth}{!}{
        \input{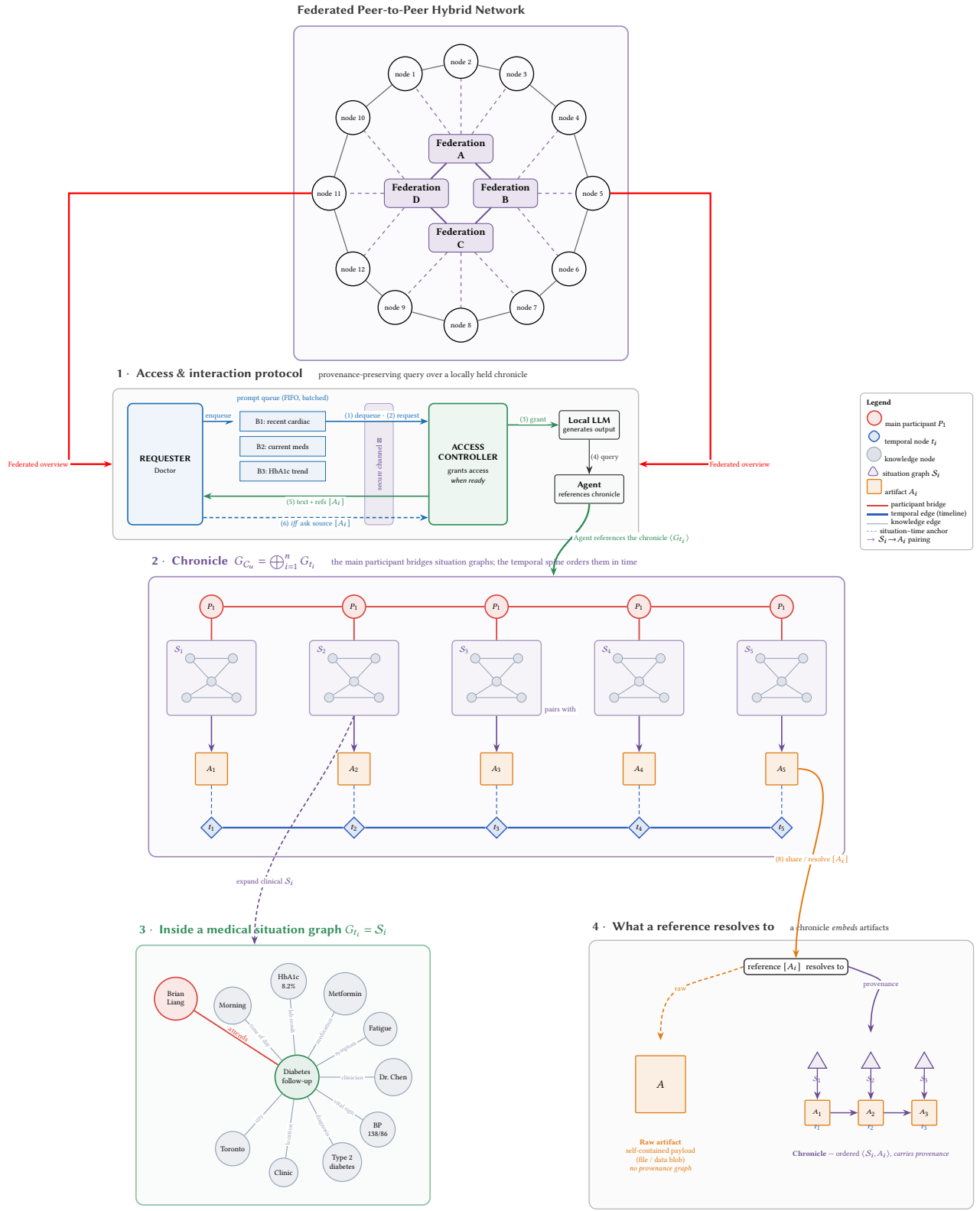}
    }
    \caption{\textbf{Federated Chronicle disclosure architecture.} A requester sends batched requests over a secure channel to an access controller, which gates a local agent and Chronicle-aware services. Holders keep temporally ordered Situation Graphs linking situations to embedded artifacts and provenance. Federated groups exchange across local networks without centralizing personal data. Responses cite provenance-carrying Chronicle entries---not isolated raw artifacts---so disclosure stays traceable and context-aware.}
    \label{fig:big}
\end{figure*}

\begin{table}[t]
\centering
\caption{Capability comparison for federated Chronicle disclosure.
\textbf{PPC}: Provenance Preserving Chronicles (our proposal).
\textbf{Hold.Sov.}: primary Chronicle custody stays with the holder, not validators or a shared ledger;
\textbf{Ledger-free}: disclosure does not require global blockchain consensus;
\textbf{Chronicle}: temporal, predicate-labeled Situation Graph semantics;
\textbf{Purpose}: exchange gated by stated request purpose;
\textbf{Relation}: access scoped to requester--holder relationship;
\textbf{Min.Sub.}: releases a minimum sufficient subgraph per request.
\yes\ = supported; \partialmark\ = partial; \no\ = absent.}
\label{tab:comparison}
\scriptsize
\setlength{\tabcolsep}{2pt}
\begin{tabular}{@{}lcccccc@{}}
\toprule
\textbf{System} & \rotatebox{60}{\textbf{Hold.Sov.}} & \rotatebox{60}{\textbf{Ledger-free}} & \rotatebox{60}{\textbf{Chronicle}} & \rotatebox{60}{\textbf{Purpose}} & \rotatebox{60}{\textbf{Relation}} & \rotatebox{60}{\textbf{Min.Sub.}} \\
\midrule
Solid              & \yes & \yes & \no  & \partialmark & \partialmark & \no  \\
MedRec             & \partialmark & \no  & \no  & \partialmark & \partialmark & \no  \\
Secret Network     & \no  & \no  & \no  & \no  & \no  & \no  \\
IPFS / libp2p      & \partialmark & \yes & \no  & \no  & \no  & \no  \\
Secure Scuttlebutt & \yes & \yes & \no  & \no  & \no  & \no  \\
\midrule
\textbf{PPC}       & \yes & \yes & \yes & \yes & \yes & \yes \\
\bottomrule
\end{tabular}
\end{table}

\section{Framework Design}
\label{sec:framework}

The core idea is simple: compile a compact authorized evidence subgraph
sufficient for the request, subject to policy---as small as policy and the
request allow, formalized as a minimization objective in \S\ref{sec:evidence}. One principle governs the
design---\emph{share no more than the request requires}---and we keep
access-control guarantees separate from what a requester's agent might infer from
the release. Figure~\ref{fig:big} illustrates the holistic architecture of our framework through a top-down hierarchical view: starting from the overarching federated peer network, zooming into the local access controller during a doctor–patient interaction, and exposing the structural anatomy of the longitudinal Chronicle—from fine-grained medical Situation Graphs down to provenance-backed artifact references

\subsection{The Authorization Function}
\label{sec:auth}

Given Chronicle $\mathcal{C}_u$ for holder~$u$, the \textbf{authorization
function}
\[
\mathcal{A}\!: \mathit{Rel} \times \mathit{Purp} \times
\bigl(\textstyle\bigcup_i V_{t_i} \cup \bigcup_i E_{t_i}\bigr)
\to \{\mathit{permit}, \mathit{deny}\}
\]
takes a requester--holder relationship $r \in \mathit{Rel}$ (e.g., treating
physician, insurance auditor), a stated purpose $p \in \mathit{Purp}$, and one
node or edge from the Chronicle, and decides: for this requester, under this
purpose, is it allowed out? Evaluation runs locally at the holder against
policies typed by a \emph{domain-expert ontology}. We write $\mathit{Rel}$
rather than a requester-identity space to keep~$\mathcal{A}$ distinct from the
request~$Q$ introduced below---authorization depends on \emph{both the requester's relationship to the holder and the operational requirements of the specific query}.

Two regulatory principles drive the function:
\begin{enumerate}[leftmargin=1.5em, itemsep=1pt, topsep=2pt]
\item \textbf{Purpose limitation:} data collected for purpose~$p_1$ may only be
disclosed for purpose~$p_2$ if $p_2$ is compatible with~$p_1$ under the
applicable legal framework (GDPR Article~5(1)(b)~\cite{gdpr2016}, HIPAA
Treatment-Payment-Operations~\cite{hipaa_min}).
\item \textbf{Relationship-based access:} the requester's relationship to the
holder---treating physician, insurance auditor, opposing counsel,
co-researcher---maps to a \emph{user type} that gates which predicate categories are in bounds. 
In PAI terms, the relationship is itself a perspective on the contextual management exchange.
\end{enumerate}

Purpose plus relationship yields a fine-grained authorization context that is
\emph{query-independent} but \emph{requester-dependent}. A treating
cardiologist asking about drug interactions can reach current prescriptions and
relevant diagnoses---not psychiatric session notes or billing records. An
insurance auditor with a different relationship and purpose gets a different
authorized slice of the same Chronicle.

\subsection{The Authorized Evidence Subgraph}
\label{sec:evidence}

Given request~$Q$, authorization context~$(r,p)$, and Chronicle
$\mathcal{C}_u = \langle G_{t_1}, \ldots, G_{t_n} \rangle$, a candidate evidence
subgraph is a pair $S = (V_S, E_S)$ with $V_S \subseteq V_{\mathcal{C}_u}$ and
$E_S \subseteq E_{\mathcal{C}_u}$ whose endpoints lie in~$V_S$, and size
$|S| = |V_S| + |E_S|$. The \textbf{authorized evidence subgraph}~$S^*$ solves the
following, where the predicate $\textsc{Sufficient}(S,Q)$ is defined
in~\S\ref{sec:sufficient-alg}:
\begin{equation}
\label{eq:optimization}
\min_{S} \; |S| \quad \text{s.t.} \;\;
\left\{
\begin{array}{ll}
S \subseteq G_{\mathcal{C}_u} & \text{(i)} \\[2pt]
\forall\, x \!\in\! V_S \cup E_S\!: \mathcal{A}(r, p, x) = \mathit{permit} & \text{(ii)} \\[2pt]
\textsc{Sufficient}(S, Q) & \text{(iii)}
\end{array}
\right.
\end{equation}

Constraint~(i): $S$ is drawn from the consolidated longitudinal Chronicle.
Constraint~(ii): every element---each node in~$V_S$ \emph{and} each edge
in~$E_S$---passes~$\mathcal{A}$, which is defined over nodes and edges alike
(\S\ref{sec:auth}).
Eq.~\eqref{eq:optimization} states the minimization objective exactly; the
protocol (\S\ref{sec:protocol}) approximates it with a tractable solver rather
than certifying a globally optimal~$S^*$, so we describe the compiled output as
\emph{compact} (minimum under the chosen objective) rather than provably
smallest.

We define \textbf{sufficiency} in two tiers; only the first is an operational
guarantee. The \emph{structural baseline} holds for any consumer: $S$ must cover
every information need with connected predicate-typed paths and satisfy
predicate-type completeness. This tier is local, decidable, and is the contract
the protocol enforces.

The second tier is an \emph{aspirational} distributional refinement---an
evaluation target, not a shipped mechanism. When the consumer is an LLM-based
agent~$\mathcal{M}$ and a holder can materialize its local \emph{authorized
subgraph} $\mathcal{C}_a = (V_a, E_a)$, with $V_a = \{v \in V_{\mathcal{C}_u} :
\mathcal{A}(r,p,v) = \mathit{permit}\}$ and $E_a$ the permitted edges
among~$V_a$ (not a bare edge set), one would like
$P_{\mathcal{M}}(\cdot \mid S, Q)$ to lie within divergence~$\varepsilon$
(measure~$D$, e.g.\ KL) of $P_{\mathcal{M}}(\cdot \mid \mathcal{C}_a, Q)$, over
\emph{semantic equivalence classes of answer strings} rather than raw tokens.
Two caveats make this a research target: (i)~in a federated deployment
$\mathcal{C}_a$ is never materialized centrally, so the check is only definable
at a single holder over its fragment; and (ii)~estimating $P_{\mathcal{M}}$ over
answer-equivalence classes is itself unsolved. The distributional
tier therefore only \emph{refines} the structural bar; when it cannot be
computed, Algorithm~\ref{alg:sufficient} falls back to the structural baseline
with no weaker guarantee.

\subsection{Sufficiency Check}
\label{sec:sufficient-alg}

\noindent\textbf{Request planning.} A planner maps~$Q$ to schema-level
\emph{information needs} $\mathcal{I}(Q) = \{i_1, \ldots, i_k\}$, each carrying a
\emph{path template} $T_j = \langle \tau_1, \ldots, \tau_m \rangle$ of predicate
types from the ontology (e.g.\ $\langle \mathit{prescribed},\\ \mathit{hasDosage}
\rangle$). Planning may be \emph{symbolic} (ontology-guided templates from a
controlled vocabulary), \emph{LLM-based} (templates proposed from
natural-language~$Q$), or \emph{hybrid} (LLM proposals validated against
ontology-admissible templates; our default). Framework soundness is conditional
on this step: sufficiency, minimization, and authorization are all defined
\emph{relative to} $\mathcal{I}(Q)$, so a planner that drops a genuine need
yields an~$S^*$ that is compact but under-informative (\S\ref{sec:medical});
auditing or recovering $\mathcal{I}(Q)$ is itself an open problem.

\smallskip\noindent\textbf{Coverage predicates.} Let $\sqsubseteq$ denote
ontology subsumption on predicate types. $\mathit{PathCov}(S, i_j)$ holds iff
$S$ contains a \emph{directed} walk
$v_0 \xrightarrow{e_1} \cdots \xrightarrow{e_m} v_m$ whose edge types match
$T_j$ up to subsumption ($\mathrm{type}(e_l) \sqsubseteq \tau_l$)---so
$\mathit{diagnosis}$ is satisfied by a $\mathit{cardiacDiagnosis}$ edge but not
conversely (subsumption-aware, not exact-label).

$\mathit{PredComplete}(S, Q)$
holds when, for every $$\tau \in \Pi(Q) = \bigcup_j \mathrm{predicateTypes}(T_j)$$,
the edges on covering walks subsuming to~$\tau$ meet the count
$\mathrm{req}(\tau, Q)$ (e.g.\ one $\mathit{hasDosage}$ per prescription).
$\mathit{Paths}(S, \mathcal{I})$ collects nodes and edges in any
$\mathit{PathCov}$ witness. Finally $\mathit{ProvComplete}(S)$ holds when every
element of~$S$ carries a provenance pointer to the Situation Graph (and artifact,
where applicable) it came from, so each released path is traceable.

\smallskip\noindent\textbf{Algorithm~\ref{alg:sufficient} is a validator}, not a
search: it decides whether a \emph{given} candidate~$S$ satisfies
constraint~(iii), and does not enumerate subgraphs, compare alternatives, or
certify minimality. That search is the compiler's job (\S\ref{sec:protocol}),
which calls the validator as a subroutine.

\begin{algorithm}[h]
\caption{Sufficiency \emph{validator} for a candidate evidence subgraph
(decides constraint~(iii); does not search or minimize)}
\label{alg:sufficient}
\small
\begin{algorithmic}[1]
\Require Candidate subgraph $S=(V_S,E_S)$; request $Q$; needs $\mathcal{I}(Q)$ with templates $\{T_j\}$;
  auth.\ context $(r,p)$; threshold $\varepsilon$;
  optional consumer model $\mathcal{M}$; local authorized view $\mathcal{C}_a$
\Ensure \textsc{true} iff $S$ is sufficient for~$Q$
\Statex
\ForAll{$x \in V_S \cup E_S$} \Comment{nodes \emph{and} edges}
  \If{$\mathcal{A}(r,p,x) = \mathit{deny}$}
    \State \Return \textsc{false} \hfill{\scriptsize(auth.\ leak)}
  \EndIf
\EndFor
\ForAll{$i_j \in \mathcal{I}(Q)$}
  \If{$\neg\mathit{PathCov}(S, i_j)$}
    \State \Return \textsc{false} \hfill{\scriptsize(unmet path template)}
  \EndIf
\EndFor
\If{$\neg\mathit{PredComplete}(S, Q)$}
  \State \Return \textsc{false} \hfill{\scriptsize(missing predicate type)}
\EndIf
\If{$\mathit{Paths}(S, \mathcal{I})$ is disconnected in $S$}
  \State \Return \textsc{false} \hfill{\scriptsize(broken linkage)}
\EndIf
\If{$\neg\mathit{ProvComplete}(S)$}
  \State \Return \textsc{false} \hfill{\scriptsize(untraceable evidence)}
\EndIf
\State $\mathit{struct} \gets \textsc{true}$
\If{$\mathcal{M}$ is LLM-based \textbf{and} $\mathcal{C}_a$ is available locally}
  \State $\pi_S \gets P_{\mathcal{M}}(\cdot \mid \mathit{serialize}(S), Q)$
  \State $\pi_a \gets P_{\mathcal{M}}(\cdot \mid \mathit{serialize}(\mathcal{C}_a), Q)$
  \State \Return $\mathit{struct} \land \bigl(D(\pi_S \,\|\, \pi_a) \le \varepsilon\bigr)$
    \hfill{\scriptsize(structural $+$ distributional)}
\Else
  \State \Return $\mathit{struct}$ \hfill{\scriptsize(structural baseline)}
\EndIf
\end{algorithmic}
\end{algorithm}

Constraint~(iii) in Eq.~\eqref{eq:optimization} holds when
Algorithm~\ref{alg:sufficient} returns \textsc{true}. The first loop rejects
any element---node or edge---failing~$\mathcal{A}$; the next four checks enforce
the mandatory \emph{structural baseline}---path coverage, predicate-type
completeness, connectivity, and provenance---for any consumer. When an LLM and
local $\mathcal{C}_a$ are available, it \emph{adds} a distributional refinement
on top; otherwise it returns the structural result alone. The validator never
selects among candidates: the compiler (\S\ref{sec:protocol}) drives the search
and invokes it to accept or reject each~$S$.

Holders run it locally on fragment~$S_i$ against
$\mathcal{I}_i \subseteq \mathcal{I}(Q)$ during pruning (\S\ref{sec:protocol});
the coordinator accepts $S^* = \bigcup_i S_i$ only when
$\bigcup_i \mathcal{I}_i = \mathcal{I}(Q)$ and each holder reports sufficiency on
its~$\mathcal{I}_i$. Since a path template can span holders, it re-runs the
algorithm once on the assembled $S^*$ against the full $\mathcal{I}(Q)$, so
cross-holder connectivity is checked, not assumed.

\section{Domain Instantiation}
\label{sec:domains}

The same protocol plugs into different regulated domains by swapping three
knobs: the domain-expert ontology, the user-type taxonomy, and the disclosure
policy. The medical case below instantiates all three and traces the protocol
from request to compiled~$S^*$.

\subsection{Medical Domain}
\label{sec:medical}

Consider patient \textsc{Alice} (holder~$u$), whose Chronicle spans fragments
at multiple hospitals. Dr.~Chen, a treating cardiologist at Hospital~A, sends an
agent request to assess drug-interaction risk before adjusting therapy. The
\emph{medical ontology} types every predicate-labeled edge into categories such
as \emph{prescription}, \emph{allergy}, \emph{diagnosis} (with subcategories
\emph{cardiac} and \emph{psychiatric}), \emph{session\_note}, and
\emph{billing}. Dr.~Chen maps to user type \emph{treating\_specialist}; the
stated purpose is \emph{drug\_interaction\_assessment}.

Request planning decomposes the natural-language ask---``Do Alice's current
medications pose interaction risk?''---into information needs
$\mathcal{I}(Q)=\{i_1,i_2\}$. Need~$i_1$ requires current prescription chains
(path template $T_1=\langle\mathit{prescribed},\mathit{hasDosage}\rangle$);
need~$i_2$ requires known allergies
($T_2=\langle\mathit{hasAllergy}\rangle$). These templates drive both
relevance retrieval and the structural sufficiency checks in
Algorithm~\ref{alg:sufficient}.

Table~\ref{tab:medical-walkthrough} summarizes the end-to-end compilation over
Alice's Hospital~A fragment~$G_{\mathit{HospA}}$. The \textbf{policy block}
(R1--R6) is what the access controller evaluates: R1 activates policy
evaluation for Dr.~Chen's relationship and purpose; R2--R3 deny entire predicate
categories irrelevant to clinical assessment (billing, psychotherapy notes);
R4 blocks psychiatric diagnoses even though other diagnoses are in scope; R5--R6
permit the prescription, allergy, and cardiac-diagnosis edges that a treating
specialist may see under this purpose. Each triple in the \textbf{Chronicle
fragment} row is tagged with its ontology category and the outcome of
$\mathcal{A}$---three edges fail (MDD diagnosis, psychiatric session note,
billing claim) before any minimization runs.

The \textbf{compilation steps} mirror Phase~1 of \S\ref{sec:protocol}.
Step~(a) applies the authorization filter, yielding authorized view
$\mathcal{C}_{a,\mathit{HospA}}$ with seven of ten edges. Step~(b) retrieves
edges matching~$i_1$ and~$i_2$. Step~(c) verifies structural sufficiency:
two full $\langle\mathit{prescribed},\mathit{hasDosage}\rangle$ walks (one
per drug), one $\langle\mathit{hasAllergy}\rangle$ walk, and predicate-type
completeness on $\Pi(Q)$. Step~(d) prunes authorized but unnecessary evidence:
the cardiac AFib diagnosis is permitted under R6 but not matched by any path
template in $\mathcal{I}(Q)$, so a purely structural minimizer drops it, keeping
situation node~$s_{\mathit{Jun24}}$ only as the minimal provenance anchor.

\smallskip\noindent\textbf{Minimality can fight utility.} That last step is
where the position gets uncomfortable, and we surface it rather than hide it in
the pruner. Dropping the AFib diagnosis is only ``minimal'' relative to a
literal reading of the request. A cardiologist assessing an amiodarone/warfarin
interaction may reasonably want the \emph{indication}---AFib is precisely why
amiodarone was prescribed and it shifts anticoagulation reasoning---so the
edge that minimization discards can be clinically load-bearing. This exposes a
real tension: aggressive minimization optimizes for disclosure safety but can
strip context a competent consumer needs, and the ``smallest sufficient''
set is only as good as the information-need decomposition $\mathcal{I}(Q)$ that
defines sufficiency. We do not claim to resolve it. It points to two design
choices a full system must make explicit---whether request planning should
expand $\mathcal{I}(Q)$ to include diagnostic \emph{indications} for retrieved
medications, and whether sufficiency should be judged against task outcome
(\S\ref{sec:evidence}) rather than path coverage alone---and it is why the
minimization target itself is an open question, not a settled default.

\smallskip\noindent\textbf{Four notions of sufficiency.} This tension is a gap
between \emph{levels} of sufficiency the formalism (\S\ref{sec:evidence}) keeps
distinct. \emph{Structural sufficiency}---path coverage, predicate-type
completeness, connectivity, provenance (Algorithm~\ref{alg:sufficient})---is the
only tier the protocol guarantees. \emph{Task sufficiency} asks whether~$S^*$
lets the consumer complete its task; \emph{clinical/legal sufficiency} adds
domain-competence and fairness norms (the AFib \emph{indication}, or discovery's
negotiation thread); and \emph{LLM-answer sufficiency} is the distributional
tier---the consumer's answer distribution over~$S^*$ tracking that over the full
authorized view~$\mathcal{C}_a$. The AFib case is structurally sufficient yet
clinically insufficient: it passes Algorithm~\ref{alg:sufficient} but omits
load-bearing context. Structural sufficiency lower-bounds the others without
implying them; closing that gap is the open question above.

The \textbf{resulting~$S^*$} under the structural policy contains six
triples---two medication chains plus one allergy edge, linked
through~$s_{\mathit{Jun24}}$. Psychiatric and billing data never cross the holder
boundary. Phase~1 delivers a grounded interaction-risk summary with provenance
references; if Dr.~Chen later needs the underlying lab image, session artifact,
or the AFib indication flagged above, Phase~2 requires Alice's explicit approval
per provenance ID.

\begin{table*}[t]
\centering
\caption{Drug-interaction request over Alice's Hospital~A Chronicle fragment
(\S\ref{sec:medical}). Denied triples marked \no; $S^*$ is the minimum released
after authorization, structural sufficiency, and pruning.}
\label{tab:medical-walkthrough}
\small
\setlength{\tabcolsep}{4pt}
\begin{minipage}{\textwidth}
\paragraph{Policy rules} (evaluated by $\mathcal{A}$ on each edge).\\
\begin{tabular}{@{}cll@{}}
\toprule
\textbf{ID} & \textbf{Condition} & \textbf{Effect} \\
\midrule
R1 & $\mathit{user\_type}=\mathit{treating\_specialist}$ $\land$
     $\mathit{purpose}=\mathit{drug\_interaction\_assessment}$ & evaluate \\
R2 & $\mathit{predicate\_category}=\mathit{billing}$ & \textbf{deny} \\
R3 & $\mathit{predicate\_category}=\mathit{session\_note}$ & \textbf{deny} \\
R4 & $\mathit{predicate\_category}=\mathit{diagnosis}$ $\land$
     $\mathit{subcategory}=\mathit{psychiatric}$ & \textbf{deny} \\
R5 & $\mathit{predicate\_category}\in\{\mathit{prescription},\mathit{allergy}\}$ & \textbf{permit} \\
R6 & $\mathit{predicate\_category}=\mathit{diagnosis}$ $\land$
     $\mathit{subcategory}=\mathit{cardiac}$ & \textbf{permit} \\
\bottomrule
\end{tabular}

\medskip
\paragraph{Chronicle fragment} $G_{\mathit{HospA}} \subset G_{\mathcal{C}_u}$
(situation $s_{\mathit{Jun24}}$ anchors the June visit).\\
\begin{tabular}{@{}lll@{}}
\toprule
\textbf{Triple $(s,p,o)$} & \textbf{Category} & \textbf{$\mathcal{A}$} \\
\midrule
$(s_{\mathit{Jun24}}, \mathit{situationOf}, \textsc{Alice})$ & context & permit \\
$(\textsc{Alice}, \mathit{hasDiagnosis}, \mathit{AFib})$ & diagnosis/cardiac & permit \\
$(\textsc{Alice}, \mathit{hasDiagnosis}, \mathit{MDD})$ & diagnosis/psychiatric & \no\ R4 \\
$(\textsc{Alice}, \mathit{prescribed}, \mathit{Warfarin})$ & prescription & permit \\
$(\mathit{Warfarin}, \mathit{hasDosage}, 5\mathit{mg})$ & prescription & permit \\
$(\textsc{Alice}, \mathit{prescribed}, \mathit{Amiodarone})$ & prescription & permit \\
$(\mathit{Amiodarone}, \mathit{hasDosage}, 200\mathit{mg})$ & prescription & permit \\
$(\textsc{Alice}, \mathit{hasAllergy}, \mathit{Penicillin})$ & allergy & permit \\
$(\textsc{Alice}, \mathit{attendedSession}, \mathit{Note447})$ & session\_note & \no\ R3 \\
$(\textsc{Alice}, \mathit{billed}, \mathit{Claim8821})$ & billing & \no\ R2 \\
\bottomrule
\end{tabular}

\medskip
\paragraph{Compilation steps.}
\textbf{(a)~Authorization filter:} remove MDD, Note447, and Claim8821;
$\mathcal{C}_{a,\mathit{HospA}}$ retains seven edges.
\textbf{(b)~Relevance retrieval:} match $i_1$ to both prescription chains and
$i_2$ to the allergy triple.
\textbf{(c)~Structural sufficiency:} two
$\langle\mathit{prescribed},\mathit{hasDosage}\rangle$ walks, one
$\langle\mathit{hasAllergy}\rangle$ walk; $\mathit{PredComplete}$ on $\Pi(Q)$.
\textbf{(d)~Minimality pruning:} drop authorized AFib diagnosis; retain
$s_{\mathit{Jun24}}$ as minimal provenance connector.

\medskip
\paragraph{Resulting $S^*$} (six triples released).\\
\begin{tabular}{@{}l@{}}
\toprule
\textbf{Released triples} \\
\midrule
$(s_{\mathit{Jun24}}, \mathit{situationOf}, \textsc{Alice})$ \\
$(\textsc{Alice}, \mathit{prescribed}, \mathit{Warfarin})$ \\
$(\mathit{Warfarin}, \mathit{hasDosage}, 5\mathit{mg})$ \\
$(\textsc{Alice}, \mathit{prescribed}, \mathit{Amiodarone})$ \\
$(\mathit{Amiodarone}, \mathit{hasDosage}, 200\mathit{mg})$ \\
$(\textsc{Alice}, \mathit{hasAllergy}, \mathit{Penicillin})$ \\
\bottomrule
\end{tabular}
\end{minipage}
\end{table*}

\subsection{Litigation Domain}

The same three knobs re-target the protocol to discovery. Opposing counsel
(user type \emph{opposing\_counsel}, purpose \emph{discovery}) requests evidence
from a corporation's Chronicle about a specific contract. The \emph{litigation
ontology} types edges as \emph{contract\_term}, \emph{communication},
\emph{internal\_memo}, \emph{financial\_record}, and
\emph{privileged} (attorney--client). Request planning decomposes ``communications
concerning Contract~X in the discovery window'' into needs
$\mathcal{I}(Q)=\{i_1,i_2\}$: $i_1$ links the contract to messages that reference
it ($T_1=\langle\mathit{governs},\mathit{referencedIn}\rangle$), $i_2$ bounds
those messages to the discovery date range ($T_2=\langle\mathit{sentOn}\rangle$).
The policy block opens contract terms and non-privileged communications inside
the window and denies attorney--client material and out-of-scope memos. Crucially,
privilege here behaves like \emph{inference closure} (\S\ref{sec:threat}): a memo
that merely quotes privileged legal advice must be denied even though its
predicate type is \emph{communication}, or the release leaks through structure.
Minimality raises the same tension as the medical case in a different guise---a
single contract clause pulled free of its surrounding negotiation thread can be
technically responsive yet misleading, so the ``smallest sufficient'' set for
\emph{fair} discovery may be larger than the one a literal template match
returns. Same protocol, different ontology, user types, and policy---and a
different~$S^*$ with the same open question about what sufficiency should mean.

\section{Protocol}
\label{sec:protocol}

The protocol runs on a federated topology: each peer keeps its Chronicle local,
a lightweight coordinator handles discovery and entity alignment, and no central
service stores or indexes raw Chronicle data. Data stays distributed; policy
gates whatever leaves the holder boundary.

Two phases implement a core invariant: \emph{text before artifacts}.

\smallskip\noindent\textbf{Phase 1: Authorized Evidence Compilation.}
\begin{enumerate}[leftmargin=1.5em, itemsep=1pt, topsep=2pt]
\item \emph{Request:} The requester's agent sends $(Q, \mathit{VP})$---request~$Q$
plus a verifiable presentation~\cite{sporny2025vc} binding identity, role, and
institutional affiliation to a decentralized identifier (DID).
\item \emph{Request planning:} A planner (symbolic or LLM-based) decomposes~$Q$
into \emph{information needs}---schema-level patterns from the domain-expert
ontology (e.g., ``current prescriptions for holder~$u$''). Holders receive
these patterns, not instance-level request content, so the ask itself does not
leak.
\item \emph{Local compilation:} Each holder~$H_i$ runs three steps on its local
fragment:
  \begin{enumerate}[label=(\alph*), leftmargin=1.2em, itemsep=0pt]
  \item \emph{Authorization filtering:} evaluate $\mathcal{A}(r,p,e)$ for each
  element, producing $\mathcal{C}_{i,a}$.
  \item \emph{Relevance retrieval:} within $\mathcal{C}_{i,a}$, pull
  Situation-Graph nodes and edges matching the information needs via entity
  linking, embedding similarity, or temporal traversal.
  \item \emph{Minimality pruning:} run subgraph optimization (e.g.,
  Prize-Collecting Steiner Tree with authorization-weighted penalties) to
  extract a compact connected subgraph covering those elements, yielding~$S_i$;
  this approximates the objective in Eq.~\eqref{eq:optimization} rather than
  guaranteeing a global minimum.
  \end{enumerate}
\item \emph{Assembly:} The coordinator merges $\{S_1, \ldots, S_k\}$ into
$S^* = \bigcup S_i$, resolves entity alignment across holders, and re-runs the
sufficiency validator (\S\ref{sec:sufficient-alg}) on the assembled~$S^*$ against
the full~$\mathcal{I}(Q)$ before delivery. Cross-holder entity alignment is a
privacy-sensitive step that the protocol \emph{depends on} but does not solve
here: we assume a privacy-preserving record-linkage primitive (e.g., Bloom-filter
matching) that links co-referent entities without revealing unmatched
identifiers, and we treat its integration with predicate-rich, temporal
Chronicle structure as an open problem.
\item \emph{Delivery:} The compiled evidence~$S^*$ is returned to the
requester's agent with \emph{provenance references} to specific Situation Graphs
and triples. When the consumer is LLM-based, $S^*$ may additionally serialize
into a grounded prompt with~$Q$; the disclosure boundary is fixed at compilation
time, not at generation time.
\end{enumerate}

This enumeration \emph{is} the compiler of Eq.~\eqref{eq:optimization}, with
Algorithm~\ref{alg:sufficient} as a subroutine: filtering~(a) enforces
constraint~(ii); retrieval~(b) and pruning~(c) \emph{search} for a small~$S$;
and the validator enforces constraint~(iii)---locally per holder on~$S_i$, then
on the merged $S^*$. The validator decides, the protocol searches; no component
both minimizes and certifies minimality, which is why we call the output
compact, not provably smallest (\S\ref{sec:evidence}).

\smallskip\noindent\textbf{Phase 2: Artifact Release (Holder-Approved).}
Phase~1 returns text plus provenance pointers---not raw artifacts. If the
requester needs the underlying asset (lab report image, audio clip, etc.), a
follow-up references the provenance IDs from Phase~1. Artifact release is
governed \emph{more} strictly than text: (i)~it re-evaluates $\mathcal{A}$ on the
artifact node and any edges it introduces, so Phase-1 authorization is necessary
but not sufficient; (ii)~it may add an artifact-release policy (higher user-type
bar or per-artifact rules) beyond the text's; and (iii)~it \emph{always} requires
explicit, per-artifact holder approval---a mandatory human gate, not a
pre-authorizable default. High-fidelity data never rides along with a text
response.

\subsection{Threat Model}
\label{sec:threat}

Two threat surfaces, handled separately; first we fix scope. \emph{In scope:} a
malicious requester exceeding authorization, inference by the consuming agent,
and prompt injection into an LLM-based \emph{consumer}. \emph{Out of scope} (as
assumptions or open problems): a \emph{coordinator} we assume honest-but-curious,
seeing only compiled, already-authorized $S_i$ and alignment metadata (a
Byzantine coordinator is future work); \emph{forged credentials}, since we
assume the verifiable-presentation/DID layer (\S\ref{sec:protocol}) is sound;
\emph{holder collusion} to reconstruct denied data; \emph{provenance-reference
leakage}, treating Phase-1 IDs as opaque capability handles that reveal nothing
without an authorized Phase-2 release (proving this is open); and \emph{prompt
injection into the planner}, a malicious~$Q$ steering $\mathcal{I}(Q)$
over-broad---a live risk, since the planner gates soundness (\S\ref{sec:evidence}).

\smallskip\noindent\textbf{Access-control threats.} A malicious requester tries
to pull Chronicle data beyond their authorization. The access controller
enforces $\mathcal{A}$ at the holder boundary; nothing leaves unless individually
permitted. We assume the holder's local system is trusted (standard in
federated learning).

\smallskip\noindent\textbf{Consumer inference threats.} Even a correct~$S^*$
can leak through inference (e.g., inferring a psychiatric diagnosis from a
psychiatric medication in~$S^*$). When the consumer is LLM-based, prompt
injection may additionally override behavioral guards~\cite{masoud2026sdrag}.
Mitigations:
\begin{enumerate}[leftmargin=1.5em, itemsep=1pt, topsep=2pt]
\item \emph{Inference closure:} optionally extend $\mathcal{A}$ so that if a
node is denied, edges that would enable high-confidence inference of that node
are denied too.
\item \emph{Decoupled disclosure:} policy applies at compilation time---not
via post-hoc redaction or instructions to self-censor.
\item \emph{Post-disclosure irreversibility:} once evidence is delivered, it
cannot be unshared. The two-phase flow limits blast radius by shipping
references first and artifacts only on explicit approval.
\end{enumerate}


\section{Conclusion}
\label{sec:conclusion}

Sharing Chronicles---the temporal, predicate-labeled graphs at the core of
Perspective-Aware AI---needs a minimum-necessary disclosure layer that today's
selective-disclosure, federated KG, and subgraph-retrieval stacks do not jointly
provide. Our technical contribution is to frame Chronicle disclosure as
\emph{minimum-necessary, policy-constrained subgraph compilation over temporal,
predicate-labeled Situation Graphs}: given a relationship, purpose, and query,
compile a compact authorized evidence subgraph---minimal under a stated
objective (Eq.~\eqref{eq:optimization}), sufficient under an explicit
provenance-complete structural check (Algorithm~\ref{alg:sufficient}), and
released provenance-first through a two-phase federated protocol. We defined the
constructs, separated the sufficiency \emph{validator} from the compiler that
searches for~$S^*$, gave an explicit threat model, and instantiated the design
in medical and litigation domains. Next up: formalize the compilation problem, ship
an efficient approximation, prototype on the PAi stack, and evaluate against a
purpose-annotated Chronicle benchmark.




\bibliographystyle{ACM-Reference-Format}
\bibliography{references}

\appendix 

\section{Open Science} 
This submission is a position and protocol-design paper. It does not
report an implemented system, empirical evaluation, newly collected
dataset, trained model, or human-subject study. Consequently, there
are no code, data, model, or experimental artifacts associated with
the claims made in the current version. All material needed to assess
the contribution---including the problem formulation, authorization
model, minimization objective, structural sufficiency conditions,
validator pseudocode, threat-model assumptions, protocol description,
and worked medical and litigation examples---is contained in the
manuscript. The examples are hypothetical and do not contain real
personal, medical, or legal records. Future implementation and
evaluation work will require a reference implementation, synthetic
Chronicle instances, machine-readable ontologies and disclosure
policies, and a purpose-annotated benchmark; these future artifacts
are not claimed or evaluated in this submission.


\section{Ethical Considerations} 

This work is motivated by reducing unnecessary disclosure of sensitive information during interactions among AI agents. Provenance Preserving Chronicles are intended to keep personal data under the holder's control, restrict disclosure according to the requester's
relationship and stated purpose, and require explicit holder approval before raw artifacts are released. 
These mechanisms may reduce exposure, but they do not eliminate ethical risk. 
An authorized subgraph can enable indirect inference of attributes that policy otherwise denies; provenance identifiers and cross-holder entity alignment can reveal metadata; and disclosed information may be copied, combined with other sources, retained indefinitely, or reused for purposes beyond the holder's expectations. 
Once information has left the holder's control, technical revocation cannot guarantee that all copies or derived inferences are removed.

Errors or bias in request planning, ontology design, authorization
policy, entity resolution, or minimization may cause either
over-disclosure or under-disclosure. Under-disclosure is particularly
important in the medical and litigation examples: an aggressively
minimized response may satisfy a structural query while omitting
context necessary for a clinically safe, legally fair, or otherwise
well-founded decision. PPC should therefore be treated as a
decision-support and disclosure-control mechanism, not as a
substitute for informed consent, professional judgment, applicable
law, or institutional governance. Consequential deployments should
use authenticated roles and purposes, least-privilege defaults,
auditable policy decisions, secure communication, privacy-preserving
entity resolution, explicit per-artifact consent, and meaningful human
review. Policies and ontologies should also be evaluated for systematic
over- or under-disclosure across demographic groups, institutions,
and resource settings. Holders should receive understandable notice
of what information was shared, with whom, for what purpose, and
the practical limits of revocation. This paper reports no
human-subject study and processes no real personal records.

\section*{Use of Generative AI}

Generative AI tools, including ChatGPT (OpenAI) and Claude (Anthropic), were used primarily
for grammar correction, language polishing, and document and LaTeX
formatting. They also assisted in preparing initial drafts of the Open
Science and Ethical Considerations sections and this disclosure
statement. The human authors determined the research problem,
technical design, formalization, threat model, analysis, and
conclusions. All AI-assisted text was critically reviewed, revised,
and verified by the authors, who retain full responsibility for the
accuracy, originality, citations, and integrity of the manuscript.

\end{document}